\documentclass[runningheads]{llncs}
\usepackage[T1]{fontenc}
\usepackage{graphicx}
\usepackage{amssymb}
\usepackage{amsmath}
\usepackage{multirow}
\usepackage{graphicx}
\usepackage{subcaption}
\begin{document}
\title{DRG-MAPPO: Hierarchical Dynamic Role-Graph Multi-Agent Reinforcement Learning for Cooperative Air Combat}
%
%
\author{Junlin Liu\inst{1, 2, 3} \and
Yang Gao\inst{1, 2}\thanks{Corresponding author} \and
Chengwei Li\inst{1, 2, 3} \and \\
Hui Chang\inst{1, 2} \and
Xinchen Zhang\inst{1, 2} \and
Zhijun Zhao\inst{1, 2} \and
Hao Zhao\inst{1, 2}
}


%
\titlerunning{Dynamic Role-Graph Multi-Agent Proximal Policy Optimization}
\authorrunning{J. Liu et al.}
%
\institute{Institute of Automation, Chinese Academy of Sciences \and
The Key Laboratory of Cognition and Decision Intelligence for Complex Systems, CASIA
\and
School of Artificial Intelligence, University of Chinese Academy of Sciences\\
\email{\{liujunlin2025, yang.gao\}@ia.ac.cn}
}
\maketitle              
\begin{abstract}
Multi-Agent Reinforcement Learning (MARL) has emerged as a pivotal paradigm for complex decision-making in autonomous systems and air combat.  
While MARL has demonstrated significant potential in air combat, achieving sophisticated tactical coordination remains a non-trivial challenge. 
This difficulty is largely attributed to two primary limitations: 
(1) the absence of structured relational modeling hinders agents from capturing complex, time-varying interactions among battlefield entities; and (2) conventional flat architectures often lack the capability to explicitly model tactical roles, leading to ambiguous task allocation in highly dynamic environments. 
To address these challenges, we propose Hierarchical \textbf{D}ynamic \textbf{R}ole-\textbf{G}raph Multi-Agent Proximal Policy Optimization (\textbf{DRG-MAPPO}), a novel MARL framework that integrates graph-based relational modeling with dynamic role assignment. 
Specifically, DRG-MAPPO constructs a graph-based representation of battlefield interactions and leverages graph attention mechanisms to extract critical relational features among allies, enemies, and threats. 
Subsequently, a high-level policy employs a dynamic role assignment mechanism to determine tactical responsibilities (e.g., ``leader'' and ``supporter''). 
Conditioned on these roles and encoded graph-relational features, a low-level policy executes discrete maneuver actions, facilitating the joint optimization of tactical strategy and collaborative execution. 
Furthermore, a target-priority auxiliary task is designed to foster the emergence of behaviors such as focus-fire. 
Experimental results demonstrate that DRG-MAPPO achieves a state-of-the-art win rate of 87\%, suggesting that our framework effectively balances relational modeling, interpretability, and optimization stability for cooperative air combat.

\end{abstract}
\section{Introduction}
Modern air combat is undergoing a fundamental paradigm shift, evolving from isolated dogfights toward sophisticated multi-UAV (Unmanned Aerial Vehicle) swarm coordination \cite{wu2025context,ding2025multi,zhu2022research,zihui2025sample,li2026evolutionary}. 
In these highly contested environments, mission success relies heavily on complex tactical synergies such as flanking maneuvers, bait-and-switch, and cooperative suppression.  
Existing cooperative air-combat decision systems often rely on expert rules \cite{jiang2018cooperative,kim2020development}, optimization approaches \cite{duan2015predator,ruan2022autonomous}, or hand-crafted tactical indicators \cite{liu2025tactical,shin2018autonomous}. These methods can encode domain-specific knowledge and behave reliably in familiar situations, but they struggle to scale in highly dynamic engagements where missile, radar, and teammate interactions jointly dictate tactical outcomes. 

Multi-Agent Reinforcement Learning (MARL) has emerged as a promising paradigm for autonomous decision-making due to its efficacy in handling high-dimensional continuous spaces \cite{kong2023hierarchical,sun2021multi,wang2024deep,li2022air,liu2026proprietary,wu2026contrastive}. 
However, existing MARL frameworks still struggle to capture the complex, time-varying topological relations among battlefield entities due to two primary limitations. (1) First, the absence of structured relational modeling impedes deep situational awareness. Without it, agents struggle to precisely perceive rapid topological shifts, thereby missing optimal tactical windows (e.g., an enemy redirecting its threat). (2) Second, traditional flat architectures lack explicit tactical role modeling, leading to role confusion and preventing the emergence of asymmetric coordination (e.g., "Leader-Wingman Decoy" tactics).

To address these challenges, we propose \textbf{D}ynamic \textbf{R}ole-\textbf{G}raph MAPPO (\textbf{DRG-MAPPO}), a hierarchical MARL framework that integrates graph-based relational modeling with dynamic role assignment. 
DRG-MAPPO decouples the decision-making process into a two-level hierarchy. 
Specifically, a graph attention mechanism first constructs structured battlefield representations to capture time-varying topological dependencies. 
Based on these representations, a high-level policy dynamically assigns explicit tactical responsibilities (e.g., ``leader'' and ``supporter'') to resolve allocation ambiguity, while a low-level policy executes precise maneuvers conditioned on these roles. 
Furthermore, we design a target-priority auxiliary task to explicitly foster asymmetric cooperative behaviors, and introduce a temporal commitment mechanism that enforces behavioral consistency within fixed decision intervals to prevent destabilizing role oscillation during training. The main contributions are summarized as follows:
\begin{itemize}
    \item We introduce DRG-MAPPO, a hierarchical MARL framework that decouples tactical role assignment from low-level maneuver control in multi-agent air combat.
    \item We design a graph-attention relational module and a target-priority auxiliary task, enabling agents to capture fluid battlefield topologies and seize optimal tactical windows.
    \item Empirical evaluations in a high-fidelity simulation environment demonstrate that DRG-MAPPO achieves an 87\% win rate, significantly outperforming state-of-the-art MARL baselines while exhibiting robust and interpretable tactical coordination.
\end{itemize}

\section{Related Work}

\subsection{Multi-Agent Reinforcement Learning in Air Combat.} 
MARL has increasingly replaced traditional rule-based and heuristic optimization methods in autonomous air combat, demonstrating significant potential in handling high-dimensional continuous state spaces \cite{wang2024deep}. 
Algorithms like MADDPG \cite{lowe2017multi}, MATD3 \cite{zhan2021twin}, and MAPPO \cite{yu2022surprising} have established a strong foundation for centralized training with decentralized execution (CTDE) in multi-agent systems.
Specifically, Wu et al. \cite{wu2025context} proposed a context-aware feature fusion method to enhance cooperative perception among multiple UAVs, improving coordination under partial observability.
Ding et al. \cite{ding2025multi} introduced a layer-delay dual-center MAPPO architecture to mitigate non-stationarity in multi-UAV decision-making. 
However, conventional MARL frameworks typically employ flat architectures that conflate strategic intent with low-level maneuver execution, lacking explicit mechanisms for modeling inter-agent topological relationships. 

\subsection{Hierarchical Reinforcement Learning (HRL).} 
To alleviate the decision-making bottleneck in long-horizon and complex tasks, HRL has been introduced into multi-agent air combat. 
By decoupling the decision process, conventional HRL frameworks typically utilize a high-level policy to generate abstract subtasks or spatial subgoals, while a low-level policy focuses on executing specific maneuver control commands to achieve these sub-objectives. 
Building on this paradigm, recent works \cite{selmonaj2023hierarchical,kong2023hierarchical2,lv2026pcsd} proposed hierarchical architectures to explicitly decouple high-level tactical decision-making from low-level maneuver control, further leveraging mechanisms like competitive self-play to optimize dual-aircraft formation engagements. 
Nevertheless, these existing hierarchical approaches often rely on heuristic state partitioning or static task assignments.

\subsection{Graph-based Relational Modeling.}
Effective coordination in multi-agent systems necessitates structured perception of inter-entity relationships. 
Graph Neural Networks (GNNs), particularly Graph Attention Networks (GAT), have been increasingly leveraged to model complex interactions and extract relational features in multi-agent decision-making. 
Jing et al. \cite{jing2024multi} employed graph convolutional networks within a MARL framework to capture the topological dependencies among machines and operations in flexible job shop scheduling. 
In the domain of UAV confrontation, Hu et al. \cite{hu2025graph} proposed a GNN-enhanced MARL approach that constructs interaction graphs among UAVs to facilitate coordinated engagement strategies. However, most existing graph-based multi-agent frameworks either limit their scope to communication topologies among friendly units or employ static graph structures, struggling to capture comprehensive, time-varying topological shifts in dense friend-or-foe engagement scenarios.


\section{Preliminaries}
\subsection{Decentralized Partially Observable Markov Decision Process}
We formulate the multi-UAV cooperative air combat problem as a Dec-POMDP, defined by the tuple $\mathcal{M} = \langle \mathcal{I}, \mathcal{S}, \{\Omega_i\}, \{\mathcal{A}_i\}, \mathcal{T}, \mathcal{Z}, \{R_i\}, \gamma \rangle$,
where $\mathcal{I} = \{1, \ldots, N\}$ is the set of $N$ cooperative agents; $\mathcal{S}$ denotes the global state space; $\Omega_i$ and $\mathcal{A}_i$ represent the local observation space and action space of each agent, with $\mathcal{A} = \prod_i \mathcal{A}_i$ being the joint action space.
The transition function $\mathcal{T}: \mathcal{S} \times \mathcal{A} \times \mathcal{S} \rightarrow [0, 1]$ governs the probability of reaching state $s_{t+1}$ from $s_t$ given joint action $a_t$.
The observation function $\mathcal{Z}: \mathcal{S} \times \mathcal{A} \rightarrow \Omega_1 \times \cdots \times \Omega_N$ maps the global state and joint action to individual observations $\{o_1, \ldots, o_N\}$ for each agent, where $o_i \in \Omega_i$.
Each agent receives a scalar reward via reward function $R_i(s_t, a_t)$, and $\gamma \in [0, 1)$ is the discount factor.
The goal is to learn a joint policy $\pi$ that maximizes the expected discounted cumulative return for each agent $i$:

  \begin{equation}
      \max_{\pi} \; \mathbb{E}_{\pi}
      \left[ \sum_{t=0}^{T} \gamma^t \, R_i(s_t, a_t) \right], \forall\, i \in \mathcal{I}
  \end{equation}

\subsection{Multi-Agent Proximal Policy Optimization (MAPPO)}
We build upon the MAPPO algorithm \cite{yu2022surprising} with a parameter-sharing scheme to enhance sample efficiency. A shared actor network $\pi_\theta$ performs decentralized execution based on local observations, while a centralized critic $V_\psi$ evaluates global states during training. Dropping the agent index $i$ for brevity, the actor minimizes a clipped surrogate loss to bound policy updates:
\begin{equation}
    \mathcal{L}_{actor}(\theta) = - \mathbb{E}_t \left[ \min \Big( \rho_t(\theta) \hat{A}_t, \; \text{clip}\big(\rho_t(\theta), 1-\epsilon, 1+\epsilon\big) \hat{A}_t \Big) \right]
\end{equation}
where $\rho_t(\theta) = \frac{\pi_\theta(a_t | o_t)}{\pi_{\theta_{\text{old}}}(a_t | o_t)}$ denotes the probability ratio between the current and the previous policy, and $\epsilon$ restricts the magnitude of policy updates. 

The objective function for the global critic aims to minimize the Mean Squared Error (MSE) between the state value estimation and the discounted empirical return $\hat{G}_t$:
\begin{equation}
    \mathcal{L}_{critic}(\psi) = \mathbb{E}_t \left[ \left( V_\psi(s_t) - \hat{G}_t \right)^2 \right]
\end{equation}
Furthermore, the advantage function $\hat{A}_t$ is calculated utilizing the Generalized Advantage Estimation (GAE) \cite{schulman2016high} to balance the variance and bias:
\begin{equation}
    \hat{A}_t = \sum_{k=0}^{\infty} (\gamma \lambda)^k \delta_{t+k},  \quad \delta_t = r_t + \gamma V_\psi(s_{t+1}) - V_\psi(s_t)
\end{equation}
where $\delta_t$ represents the standard TD error, $\gamma$ is the discount factor, and $\lambda$ acts as the GAE smoothing parameter.

\section{Simulation Environment}

\subsection{Observation Space} \label{sec:obs}
\begin{figure}[t]
    \centering
    \includegraphics[width=0.99\textwidth]{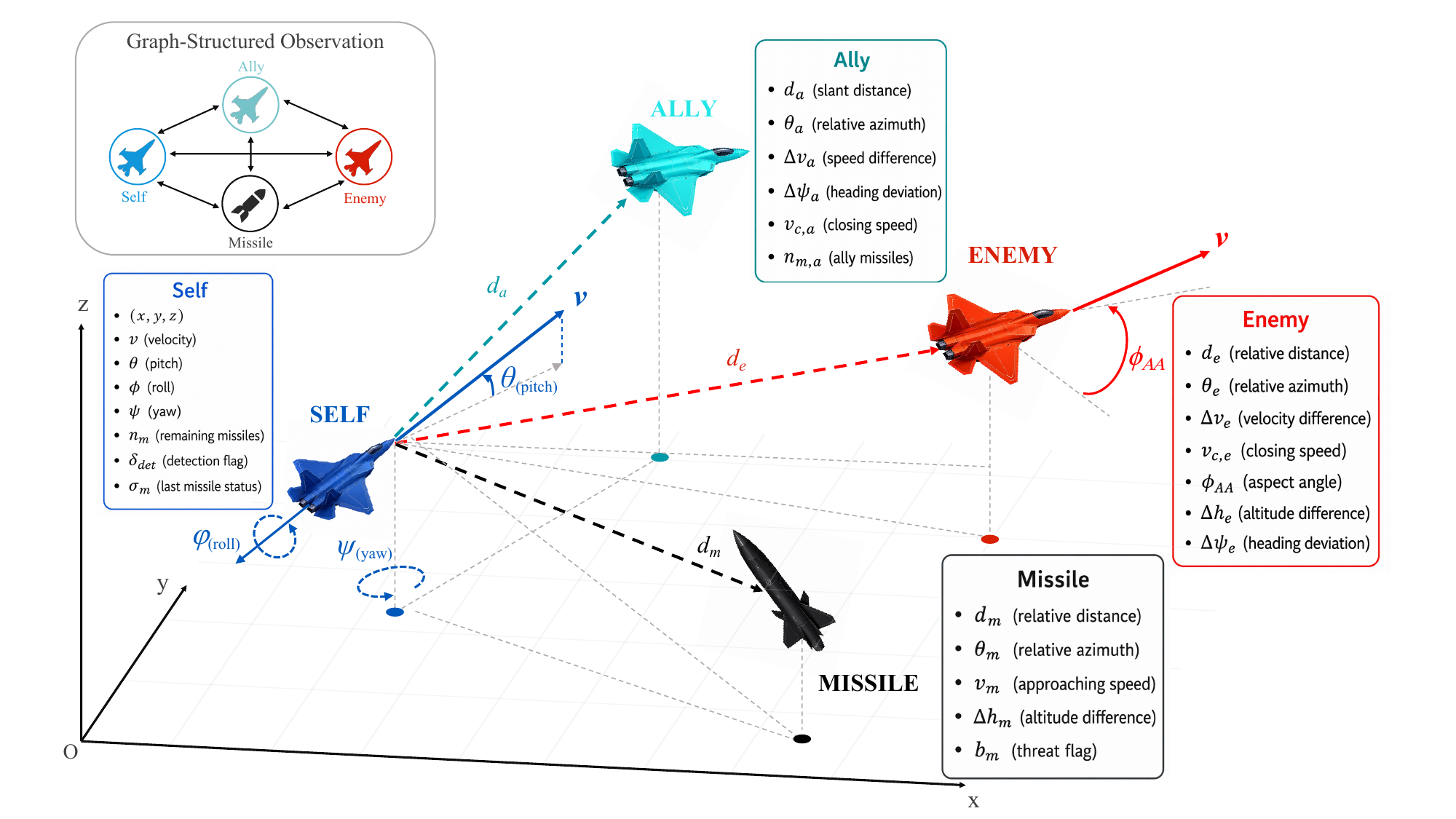} 
    \caption{Overview of the battlefield tactical geometry and observation space.}
    \label{fig:observation}
\end{figure}

Unlike conventional monolithic state vectors, we model the battlefield at each step $t$ as a directed graph $\mathcal{G}_t = (\mathcal{V}_t, \mathcal{E}_t)$ to explicitly capture entity interactions. 
As illustrated in Fig.~\ref{fig:observation}, the node set is partitioned into four categories ($\mathcal{V}_t = \mathcal{V}_{\text{self}} \cup \mathcal{V}_{\text{ally}} \cup \mathcal{V}_{\text{enemy}} \cup \mathcal{V}_{\text{missile}}$), where the self-node encodes the agent's intrinsic state and the remaining nodes encode relative features. The exact mathematical compositions and detailed definitions of all feature components are summarized in Table~\ref{tab:obs}.

\textbf{Self-node features.} The self-state $\mathbf{f}_{\text{self}} \in \mathbb{R}^{10}$ captures the agent's kinematic state and resource availability:
  \begin{equation}
      \mathbf{f}_{\text{self}} = [x, y, z, v, \theta, \phi, \psi, n_m, \delta_{\text{det}}, \sigma_m]
  \end{equation}
   
\textbf{Ally-node features.} For each ally $j \in \mathcal{V}_{\text{ally}}$, the relative feature vector $\mathbf{f}_{\text{ally}}^{j} \in \mathbb{R}^{6}$ is defined as:
  \begin{equation}
      \mathbf{f}_{\text{ally}}^{j} = [d_a^j, \theta_a^j, \Delta v_a^j, \Delta \psi_a^j, v_{c,a}^j, n_{m,a}^j]
  \end{equation}
  
\textbf{Enemy-node features.} For each detected enemy $k \in \mathcal{V}_{\text{enemy}}$, the feature vector $\mathbf{f}_{\text{enemy}}^{k} \in \mathbb{R}^{7}$ encodes the tactical geometry:
  \begin{equation}
      \mathbf{f}_{\text{enemy}}^{k} = [d_e^k, \theta_e^k, \Delta v_e^k, v_{c,e}^k, \phi_{\text{AA}}^k, \Delta h_e^k, \Delta \psi_e^k]
  \end{equation}
   
\textbf{Missile-node features.} For each incoming missile $l \in \mathcal{V}_{\text{missile}}$, the threat feature vector $\mathbf{f}_{\text{missile}}^{l} \in \mathbb{R}^{5}$ is defined as:
  \begin{equation}
      \mathbf{f}_{\text{missile}}^{l} = [d_m^l, \theta_m^l, v_m^l, \Delta h_m^l, b_m^l]
  \end{equation}

\begin{table*}[t] 
  \centering
  \caption{Summary of state variables in the graph-structured observation.}\label{tab:obs}
  
  \resizebox{\textwidth}{!}{
      \begin{tabular}{c l l @{\hspace{3em}} c l l}
      \hline
      \textbf{Type} & \textbf{Symbol} & \textbf{Description} & \textbf{Type} & \textbf{Symbol} & \textbf{Description} \\
      \hline
      \multirow{8}{*}{Self}
      & $x, y, z$ & Position coordinates & \multirow{7}{*}{Enemy} & $d_e$ & Relative distance \\
      & $v$ & Velocity magnitude & & $\theta_e$ & Relative azimuth angle \\
      & $\theta$ & Pitch angle & & $\Delta v_e$ & Velocity difference \\
      & $\phi$ & Roll angle & & $v_{c,e}$ & Closing speed \\
      & $\psi$ & Yaw angle & & $\phi_{\text{AA}}$ & Aspect angle \\
      & $n_m$ & Remaining missiles & & $\Delta h_e$ & Altitude difference \\
      & $\delta_{\text{det}}$ & Detected by enemy (binary) & & $\Delta \psi_e$ & Heading deviation \\
      \cline{4-6} 
      
      & $\sigma_m$ & Last missile status & \multirow{6}{*}{Ally} & $d_a$ & Slant distance \\
      \cline{1-3} 
      
      \multirow{5}{*}{Missile}
      & $d_m$ & Relative distance & & $\theta_a$ & Relative azimuth angle \\
      & $\theta_m$ & Relative azimuth angle & & $\Delta v_a$ & Airspeed difference \\
      & $v_m$ & Approaching speed & & $\Delta \psi_a$ & Heading deviation \\
      & $\Delta h_m$ & Altitude difference & & $v_{c,a}$ & Closing speed \\
      & $b_m$ & Threat existence flag (binary) & & $n_{m,a}$ & Ally's remaining missiles \\
      \hline
      \end{tabular}
  }
\end{table*}

\subsection{Action Space}
To reduce exploration costs, we eschew continuous or fine-grained control in favor of a tactical-level discrete action space. The low-level policy outputs a discrete command $a_i^t \in \mathcal{A}$, which an underlying autopilot translates into flight trajectories, decoupling strategic decisions from low-level execution. The action set $\mathcal{A} = \mathcal{A}_{\text{maneuver}} \cup \mathcal{A}_{\text{weapon}}$ comprises 12 commands covering tactical maneuvering (adjustment, pursuit, evasion) and missile engagement, as detailed in Table~\ref{tab:action}.


\begin{table*}[ht]
  \centering
  \caption{Summary of the tactical discrete action space for each agent.}\label{tab:action}

  \resizebox{\textwidth}{!}{
      \begin{tabular}{c l l @{\hspace{3em}} c l l}
      \hline
      \textbf{Index} & \textbf{Category} & \textbf{Action Description} & \textbf{Index} & \textbf{Category} & \textbf{Action Description}
   \\
      \hline
      0 & \multirow{6}{*}{Maneuver} & Maintain current heading and altitude & 6 & \multirow{5}{*}{Maneuver} & Dive to decrease altitude
  \\
      1 & & Turn left by 30$^\circ$ toward target & 7 & & Accelerate to gain closing speed \\
      2 & & Turn left by 60$^\circ$ toward target & 8 & & Decelerate to reduce closing speed \\
      3 & & Turn right by 30$^\circ$ toward target & 9 & & Perform S-shaped lateral oscillation \\
      4 & & Turn right by 60$^\circ$ toward target & 10 & & Execute notch maneuver to evade missile \\
      \cline{4-6}
      5 & & Climb to increase altitude & 11 & Weapon & Launch missile at the designated target \\
      \hline
      \end{tabular}
  }
  \end{table*}

\subsection{Reward Function}
  To address the sparse reward inherent in air combat, we design a composite reward combining a shared team signal with individual dense
  shaping terms:
  \begin{equation}\label{eq:reward}
      r_t^i = r_{\text{combat}} + \alpha \, r_{\text{adv},t}^{i} + \beta \, r_{\text{threat},t}^{i} + r_{\text{bound},t}^{i}
  \end{equation}

  The team-shared combat reward $r_{\text{combat}}$ provides $+R_{\text{win}}$ / $-R_{\text{lose}}$ at episode termination, and
  $+r_{\text{kill}}$ / $-r_{\text{death}}$ upon individual destruction events.
  The tactical advantage reward offers dense offensive guidance based on the Antenna Train Angle (ATA) and relative distance:
  \begin{equation}\label{eq:adv}
      r_{\text{adv},t}^{i} = \left(1 - \frac{\theta_{\text{ATA}}^{i}}{\pi}\right) \cdot \exp\left(-\frac{d_e^{i}}{D_{\max}}\right)
  \end{equation}
  where $\theta_{\text{ATA}}^{i} \in [0, \pi]$ is the angle between agent $i$'s velocity vector and the line-of-sight to its nearest
  enemy, and $D_{\max}$ is the maximum engagement range. This term is maximized when the agent's nose aligns with a nearby enemy.

  The threat avoidance penalty compels defensive maneuvers when an incoming missile is detected:
  \begin{equation}\label{eq:threat}
      r_{\text{threat},t}^{i} = -1(b_m^{i} = 1) \cdot \exp\left(-\frac{d_m^{i}}{D_{\text{safe}}}\right)
  \end{equation}
  where $D_{\text{safe}}$ is the safety distance threshold.

  A fixed boundary penalty is imposed when agents depart the predefined operational zone $\mathcal{B}$:
  \begin{equation}\label{eq:bound}
      r_{\text{bound},t}^{i} = -R_{\text{oob}} \cdot 1(p_t^{i} \notin \mathcal{B})
  \end{equation}
  where $p_t^{i}$ is the position of agent $i$ and $R_{\text{oob}}$ is the out-of-boundary penalty constant.
  Notably, cooperative behaviors (e.g., focus-fire) are not explicitly rewarded but instead emerge from the synergy of the graph
  attention module, dynamic role assignment, and the target-priority auxiliary task, avoiding reward over-engineering.

\section{Methods}
\subsection{DRG-MAPPO Framework}
\begin{figure}[t]
    \centering
    \includegraphics[width=0.99\textwidth]{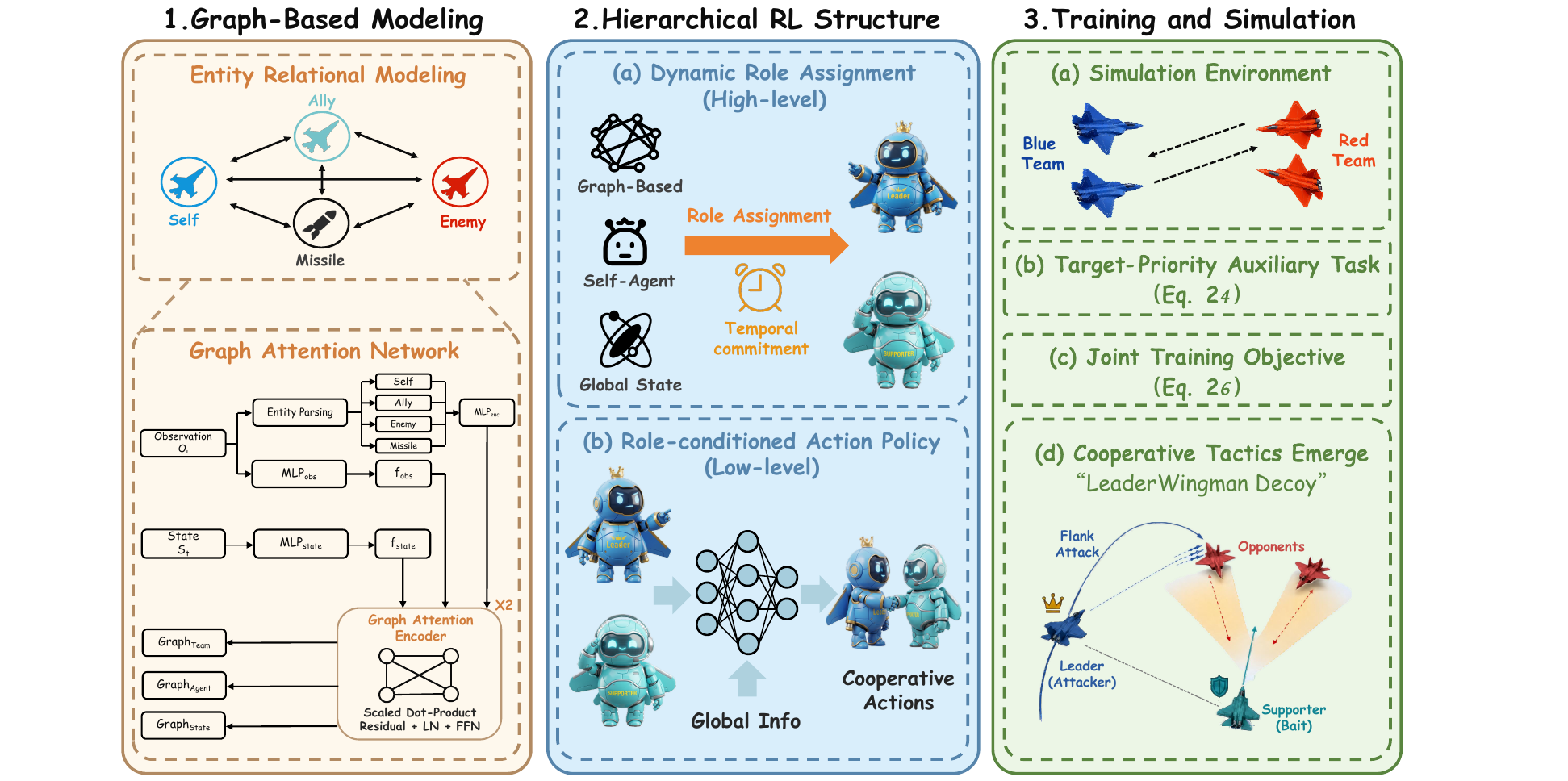} 
    \caption{Overview of the DRG-MAPPO framework. (1) graph-based state modeling for entity relationship extraction, (2) a hierarchical RL structure for tactical decision-making, and (3) centralized training and decentralized simulation.}
    \label{fig:framework}
\end{figure}

The DRG-MAPPO framework (Fig.~\ref{fig:framework}) integrates three core components: (1) a graph-based relational encoder for capturing time-varying topological interactions among entities; (2) a hierarchical policy decoupling high-level tactical role assignment from low-level role-conditioned maneuvers; and (3) a target-priority auxiliary task injecting domain inductive biases to foster cooperation. The framework is optimized end-to-end via the CTDE paradigm, where the centralized critic leverages global states during training to stabilize optimization, while actors execute relying strictly on local observations.

\subsection{Graph-Based Entity Relational Modeling}

  \textbf{Entity graph construction.}
  Rather than treating the raw observation as a monolithic feature vector, we
  decompose it into a local entity graph $\mathcal{G}_i = (\mathcal{V}_i,
  \mathcal{E}_i)$ for each agent $i$. The node set $\mathcal{V}_i$ consists of four entity categories as defined in the observation space (Section~\ref{sec:obs}): self, ally, enemy, and missile.
  Each node $k \in \mathcal{V}_i$ is characterized by its raw feature vector
  $\mathbf{x}_k$ extracted from the corresponding observation partition.
  Since entity types possess heterogeneous feature dimensions, all node features
   are zero-padded to a uniform dimension $d_{\text{raw}}$ and augmented with a
  learned type embedding $\mathbf{e}_{\text{type}(k)} \in \mathbb{R}^{d_t}$ that
   encodes entity category information. 
   The combined features are projected into a shared hidden space via a two-layer
   MLP, $d_h$ is the hidden dimension and $[\,;\,]$ denotes concatenation:
  \begin{equation}\label{eq:node_enc}
      h_k = \text{MLP}\left(
        [\text{pad}(x_k) \,;\, e_{\text{type}(k)}] \right), \quad
        h_k \in \mathbb{R}^{d_h}
  \end{equation}

\textbf{Graph attention encoder.}
To capture the relational structure among entities, we implement a graph attention mechanism using a scaled dot-product formulation \cite{vaswani2017attention}. This approach operates over a fully-connected entity graph and was empirically found to provide more stable training dynamics, particularly for the small-scale graphs characteristic of our environment.
Given the initial node representations $H^{(0)} =
  [h_1^{(0)}, \ldots, h_{|\mathcal{V}_i|}^{(0)}]^{\top} \in
  \mathbb{R}^{|\mathcal{V}_i| \times d_h}$, the attention-based relational
  update is computed as:
  \begin{equation}\label{eq:qkv}
      Q = H^{(0)} \mathbf{W}_Q, \quad K =
  H^{(0)} \mathbf{W}_K, \quad V = H^{(0)}
  \mathbf{W}_V
  \end{equation}
  \begin{equation}\label{eq:attn}
      A = \text{softmax}\left(
  \frac{QK^{\top}}{\sqrt{d_h}} \right)
  \end{equation}
  \begin{equation}\label{eq:graph_update}
      \tilde{H} = \text{LayerNorm}\left( H^{(0)} + AV \right), \quad
      H^{(1)} = \text{LayerNorm}\left( \tilde{H} +
  \text{FFN}(\tilde{H}) \right)
  \end{equation}
where $\mathbf{W}_Q, \mathbf{W}_K, \mathbf{W}_V \in \mathbb{R}^{d_h \times d_h}$ are learnable projection matrices, and $\text{FFN}(\cdot)$ denotes a two-layer feed-forward network with ReLU activation. The attention matrix $A \in \mathbb{R}^{|\mathcal{V}_i| \times |\mathcal{V}_i|}$ captures pairwise relational importance: entry $A_{jk}$ reflects how much entity $j$ attends to entity $k$. 
To facilitate multi-level relational reasoning, we derive three levels of representation from the output $H^{(L)}$ of the final graph attention layer:
  \begin{equation}\label{eq:representations}
      z_{\text{self}}^{i} = H^{(L)}[0], \quad
      g_{\text{agent}}^{i} = \frac{1}{|\mathcal{V}_i|} \sum_{k} H^{(L)}[k], \quad
      g_{\text{team}} = \frac{1}{N} \sum_{i=1}^{N}
  g_{\text{agent}}^{i}
  \end{equation}
  where $z_{\text{self}}^{i} \in \mathbb{R}^{d_h}$ is the graph-refined
   self-representation; $g_{\text{agent}}^{i} \in \mathbb{R}^{d_h}$ summarizes agent $i$'s local battlefield topology; and $g_{\text{team}} \in
  \mathbb{R}^{d_h}$ provides a team-level abstraction shared across agents.


\subsection{Hierarchical Policy with Dynamic Role Assignment}

  \textbf{Role policy (high-level).}
  The high-level role policy assigns a discrete tactical role $r_i \in \{1,
  \ldots, K\}$ to each agent, enabling explicit division of labor. The role
  policy takes as input the multi-level relational representations and the
  global context:
  \begin{equation}\label{eq:global_context}
      f_{\text{state}} = \text{MLP}_{\text{state}}(s), \quad
      c_{\text{global}} =
  \text{MLP}_{\text{fuse}}([f_{\text{state}} \,;\, g_{\text{team}}])
  \end{equation}
  \begin{equation}\label{eq:role_policy}
      \pi_{\text{role}}(r_i \mid o_i) = \text{softmax}\left(
  \text{MLP}_{\text{role}}\left( [z_{\text{self}}^{i} \,;\,
  g_{\text{agent}}^{i} \,;\, c_{\text{global}}] \right)
  \right)
  \end{equation}
  The selected role index is mapped to a continuous representation through a
  learned role embedding table $E_{\text{role}} \in \mathbb{R}^{K
  \times d_r}$. Unlike one-hot encodings, the learned embedding allows the framework to
  capture latent similarities between roles and provides richer gradient signals
   to downstream policy layers.
  \begin{equation}\label{eq:role_emb}
      e_{\text{role}}^{i} = E_{\text{role}}[r_i], \quad
      e_{\text{role}}^{i} \in \mathbb{R}^{d_r}
  \end{equation}

  \textbf{Temporal commitment mechanism.}
  In air combat, tactical roles correspond to sustained behavioral modes (e.g.,
  maintaining offensive pursuit or providing suppressive cover) rather than
  frame-by-frame switching. To reflect this and prevent destabilizing role
  oscillation, we introduce a temporal commitment mechanism that re-samples
  roles only at fixed intervals:
  \begin{equation}\label{eq:temporal}
      r_i^t =
      \begin{cases}
          r_i^{t} \sim \pi_{\text{role}}(\cdot \mid o_i^t) & \text{if } t\mod T_{\text{role}} = 0 \\
          r_i^{t-1} & \text{otherwise}
      \end{cases}
  \end{equation}
  where $T_{\text{role}}$ is the commitment horizon. This mechanism provides
  three benefits: (i) it enforces behavioral consistency within each commitment
  window; (ii) it enables teammates to anticipate each other's sustained
  intentions, facilitating implicit coordination; and (iii) it reduces the
  effective decision frequency of the high-level policy, stabilizing training.

  \textbf{Role-conditioned action policy (low-level).}
  Conditioned on the assigned role embedding, the low-level policy selects a
  discrete action from the tactical action space:
  \begin{equation}\label{eq:actor}
      \pi_{\theta}(a_i \mid o_i, r_i) = \text{softmax}\left(
  \text{MLP}_{\text{actor}}\left( [f_{\text{obs}}^{i} \,;\,
  z_{\text{self}}^{i} \,;\, g_{\text{agent}}^{i} \,;\,
  c_{\text{global}} \,;\, e_{\text{role}}^{i} \,;\,
  \text{id}_i] \right) \right)
  \end{equation}
  where $\text{id}_i$ is a one-hot agent identity vector that enables parameter
  sharing while allowing agent-specific behaviors. The centralized critic
  estimates the state-value function using global information:
  \begin{equation}\label{eq:critic}
      V_{\psi}(s, r_i) = \text{MLP}_{\text{critic}}\left(
  [f_{\text{state}} \,;\, z_{\text{self}}^{i} \,;\,
  c_{\text{global}} \,;\, e_{\text{role}}^{i} \,;\,
  \text{id}_i] \right)
  \end{equation}
  This design adheres to the CTDE principle: the actor (Eq.~\ref{eq:actor}) uses only local features for decentralized execution, while the critic
   (Eq.~\ref{eq:critic}) accesses the global state $s$ through
  $f_{\text{state}}$ during centralized training. Both are conditioned
  on the role embedding, ensuring that the value estimation accounts for the
  agent's current tactical responsibility.

\subsection{Target-Priority Auxiliary Task}To foster cooperative behaviors like focus-fire without complex reward shaping, we introduce an auxiliary target-prioritization task that provides dense supervisory signals. As shown in Eq.~\eqref{eq:aux_head}, the auxiliary head maps concatenated self-features $z_{\text{self}}^{i}$, graph-derived features $g_{\text{agent}}^{i}$, and role embeddings $e_{\text{role}}^{i}$ to a probability distribution $p_{\text{aux}}^{i} \in \mathbb{R}^{N_e}$ over $N_e$ enemy targets:\begin{equation}\label{eq:aux_head}p_{\text{aux}}^{i} = \text{softmax}\left( \text{MLP}_{\text{aux}}\left( [z_{\text{self}}^{i};  g_{\text{agent}}^{i} ; e_{\text{role}}^{i}] \right) \right)\end{equation}Ground-truth labels $y_{\text{target}}$ are derived from a heuristic priority scoring function:\begin{equation}\label{eq:target_score}\text{score}(i, j) = d_{ij} - \alpha_1 \cdot L_{\text{own}}^{ij} + \alpha_2 \cdot L_{\text{enemy}}^{ij} - \alpha_3 \cdot M_j\end{equation}where $d_{ij}$ is the distance to enemy $j$; $L_{\text{own}}^{ij}, L_{\text{enemy}}^{ij} \in \{0, 1\}$ denote radar lock statuses; and $M_j$ is the number of friendly missiles already tracking enemy $j$. The enemy with the lowest score is designated as the primary target. Crucially, the auxiliary head shares the graph attention encoder with the main policy and is optimized via cross-entropy (CE) loss: $\mathcal{L}_{\text{aux}} = \text{CE}(p_{\text{aux}}, y_{\text{target}})$.

\subsection{Joint Training Objective}
The framework is trained end-to-end via the Adam optimizer. The joint objective combines PPO clipped surrogates with auxiliary and entropy terms:
\begin{equation}\label{eq:joint_loss}
\mathcal{L} = \mathcal{L}_{\text{actor}} + \lambda_r \mathcal{L}_{\text{role}} + \lambda_V \mathcal{L}_{\text{critic}} + \lambda{_\text{aux}} \mathcal{L}_{\text{aux}} - \sum \lambda_H H(\pi)
\end{equation}
where $\mathcal{L}_{\text{actor}}$ and $\mathcal{L}_{\text{role}}$ denote clipped losses for action and role policies; $\mathcal{L}_{\text{critic}}$ and $\mathcal{L}_{\text{aux}}$ represent the value MSE and auxiliary CE loss, respectively. $H(\pi)$ denotes entropy bonuses for both policy levels. Crucially, role gradients are computed only at re-sampling intervals ($t \mod T_{\text{role}} = 0$), using a distinct clipping coefficient $\epsilon_r$ to accommodate the lower decision frequency of role assignments without undermining temporal commitment stability.

\section{Experiments and Results}
\subsection{Simulation Setting}
Our 2v2 Beyond-Visual-Range (BVR) engagement is modeled within a 200 km $\times$ 100 km operational theater. The scenario initializes with Blue and Red flights (two aircraft each) deployed at opposing southern and northern boundaries. Every combatant is equipped with four air-to-air missiles and maintains a standard combat profile: an altitude of 3 km and an initial cruise speed of 180 m/s. The simulation environment, developed in C++ and visualized via GTacview, enforces strict termination criteria: an episode ends upon total team elimination, boundary breach, or exceeding the maximum time limit. Victory is determined by numerical superiority at termination, while identical survival counts result in a draw. The maximum time limit is set to 800 seconds to ensure sufficient duration for tactical engagement and missile fly-out.

\subsection{Main Results and Performance Analysis}

\begin{figure}[t]
  \centering
  \begin{subfigure}[b]{0.49\textwidth}
      \centering
      \includegraphics[width=\textwidth]{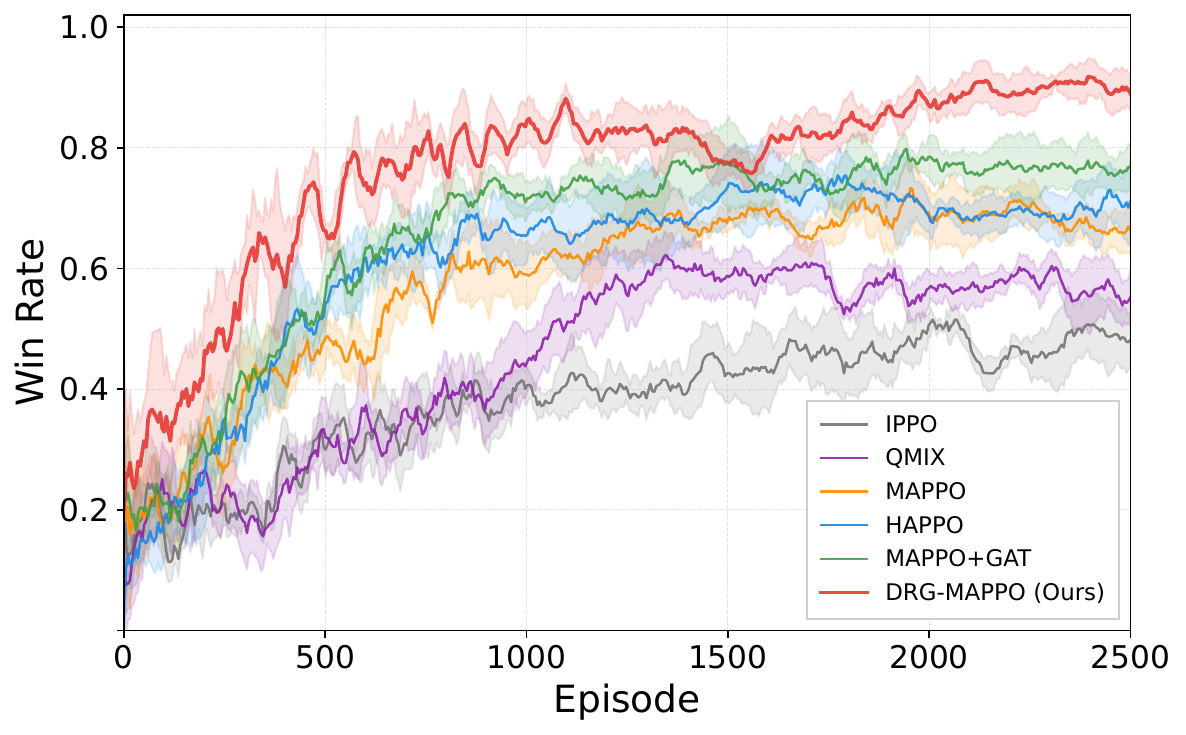}
      \caption{Win rate curves}
      \label{fig:winrate}
  \end{subfigure}
  \hfill
  \begin{subfigure}[b]{0.49\textwidth}
      \centering
      \includegraphics[width=\textwidth]{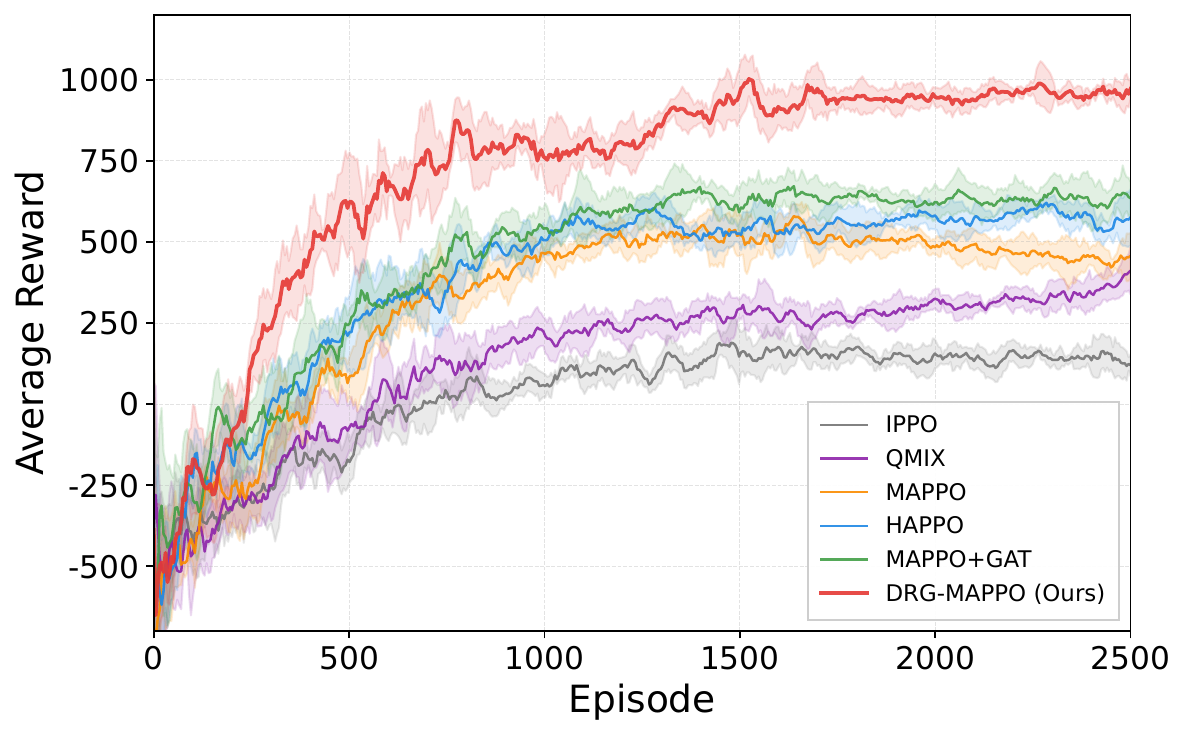}
      \caption{Average reward curves}
      \label{fig:reward}
  \end{subfigure}
  \caption{Training curves of DRG-MAPPO and other baselines in the 2v2 scenario.}
  \label{fig:training_curves}
  \end{figure}

To evaluate the effectiveness of DRG-MAPPO, we present the training curves of all methods over 2500 episodes in Fig.~\ref{fig:training_curves}.

  \textbf{Win Rate.} During the initial exploration phase (0--800 episodes), all methods exhibit a similar cold-start bottleneck, reflecting the inherent difficulty of learning complex 2v2 coordination from high-dimensional observations. After approximately 1250 episodes, DRG-MAPPO begins to significantly outperform the baselines, ultimately achieving a peak win rate of 87\%. This performance gap underscores the synergy between relational modeling and hierarchical role assignment. While MAPPO+GAT and HAPPO yield moderate improvements by incorporating graph structures or hierarchical decomposition alone, they fall short of the integrated DRG-MAPPO architecture. In contrast, standard MAPPO lacks the structural inductive biases necessary for complex entity reasoning. Furthermore, the inferior performance of QMIX and IPPO highlights that value decomposition and independent learning are insufficient for the tight tactical coordination required in adversarial BVR engagements. This demonstrates that decoupling macro-level tactical role assignment from micro-level decision-making is vital for overcoming the exploration bottleneck and achieving robust multi-agent coordination in air combat.
  

  \textbf{Average Reward.} 
  The evolution of the average episodic reward, illustrated in Fig.~\ref{fig:reward}, further corroborates the performance gains. During the initial exploration phase, all agents receive similar negative rewards (approximately $-450$), primarily due to frequent boundary breaches and missile attrition. As basic engagement behaviors emerge, the curves surpass the zero-threshold around episode 400. DRG-MAPPO ultimately converges to the highest asymptotic reward ($\sim 950$), outperforming the mid-tier methods, which plateau between 500 and 650. Consistent with findings in similar BVR benchmarks, standard MAPPO often struggles with exploration efficiency in discrete spaces, while independent learners like IPPO exhibit the lowest rewards due to severe non-stationarity.


\subsection{Ablation Experiments}
\begin{figure}[t]
  \centering
  \begin{subfigure}[b]{0.49\textwidth}
      \centering
      \includegraphics[width=\textwidth]{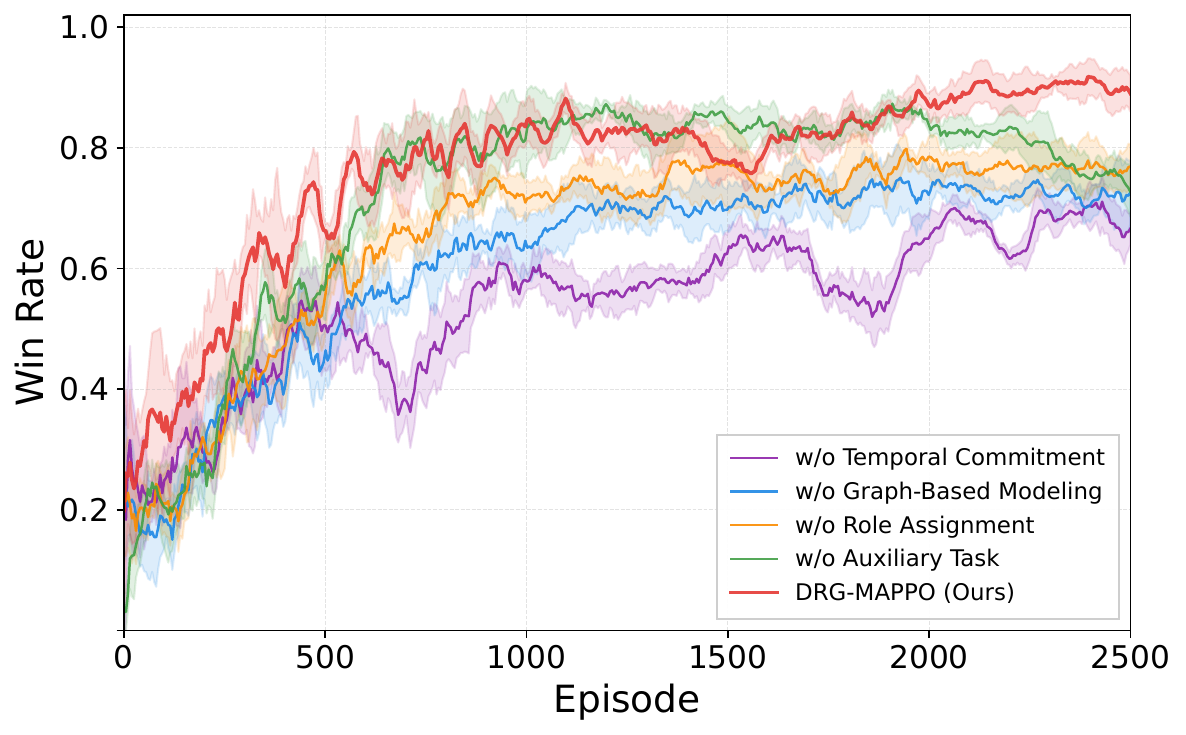}
      \caption{Ablation study curves}
      \label{fig:ablation}
  \end{subfigure}
  \hfill
  \begin{subfigure}[b]{0.4\textwidth}
      \centering
      \includegraphics[width=\textwidth]{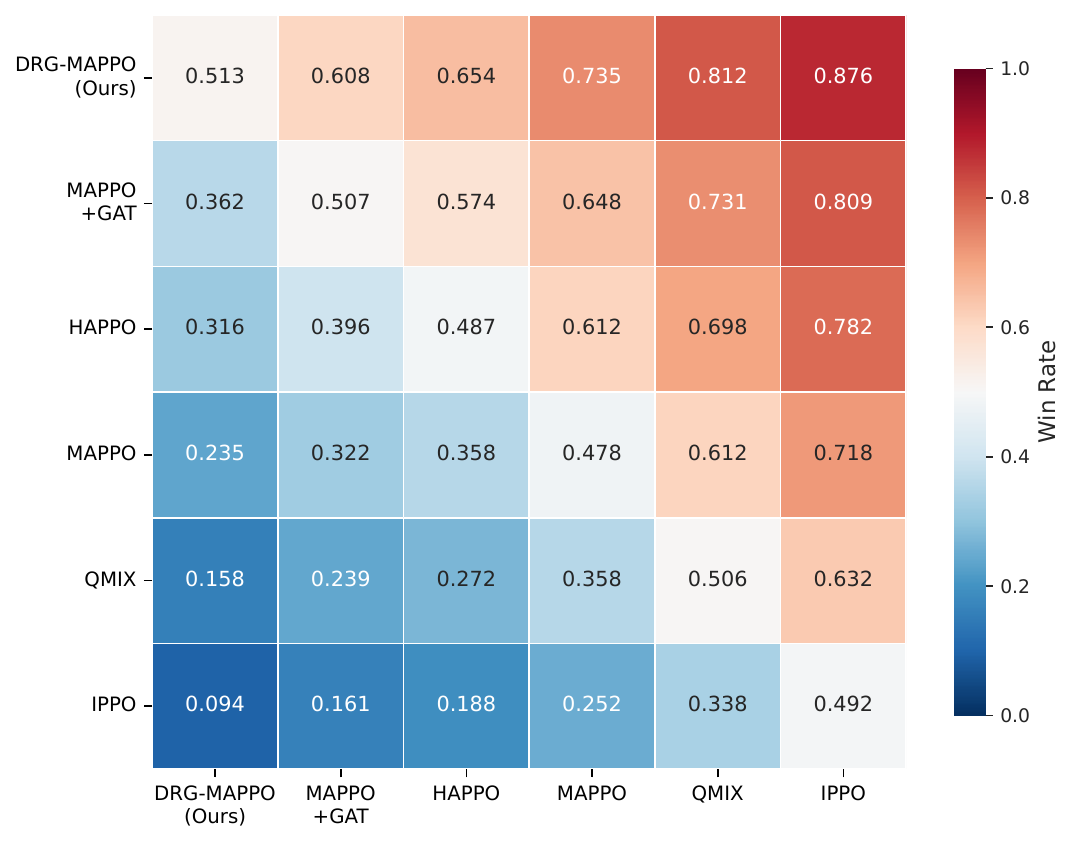}
      \caption{The heatmap of experiments}
      \label{fig:crossplay}
  \end{subfigure}
  \caption{Comprehensive evaluation in the 2v2 scenario. (a) The training win rates of ablation experiments. (b) The heatmap of confrontation experiments.}
  \label{fig:ablation_and_heatmap}
  \end{figure}

To quantify the individual contribution of each component, we evaluate several ablation variants of DRG-MAPPO. 
As illustrated in Fig.~\ref{fig:ablation}, the superior performance and tactical efficiency of DRG-MAPPO stem from the synergy of four core modules: (1) Graph-Based Modeling facilitates structured relational reasoning; (2) Dynamic Role Assignment fosters emergent division of labor; (3) Target-priority Auxiliary Task effectively shapes entity representations toward combat-critical features; and (4) Temporal Commitment prevents high-frequency role oscillations that would otherwise lead to tactical inconsistency and severe performance fluctuations.

\subsection{Confrontation Experiments}
To further evaluate the generalization and robustness of the learned policies, we conduct a round-robin cross-play tournament among all models. Each pair of policies competes for 200 independent episodes; the resulting win-rate matrix is visualized in Fig.~\ref{fig:crossplay}. 
As shown in the matrix, DRG-MAPPO achieves a superior average win rate of 69.9\%, consistently outperforming all baselines in head-to-head matchups. This performance advantage demonstrates that our role-based coordination strategy generalizes effectively to diverse adversarial behaviors instead of overfitting to the specific patterns of a single training opponent. Furthermore, the minor deviations from 0.5 observed on the diagonal are attributed to the intrinsic stochasticity in the simulation environment.

\subsection{Tactical Behavior Visualization}
To qualitatively validate the interpretability and coordination of DRG-MAPPO, we visualize a 2v2 engagement episode. The trained DRG-MAPPO model controls the blue-side agents, while the red side employs a rule-based policy. 

\begin{figure*}[t]
  \centering
  \begin{subfigure}[b]{0.28\textwidth}
      \centering
      \includegraphics[width=\textwidth]{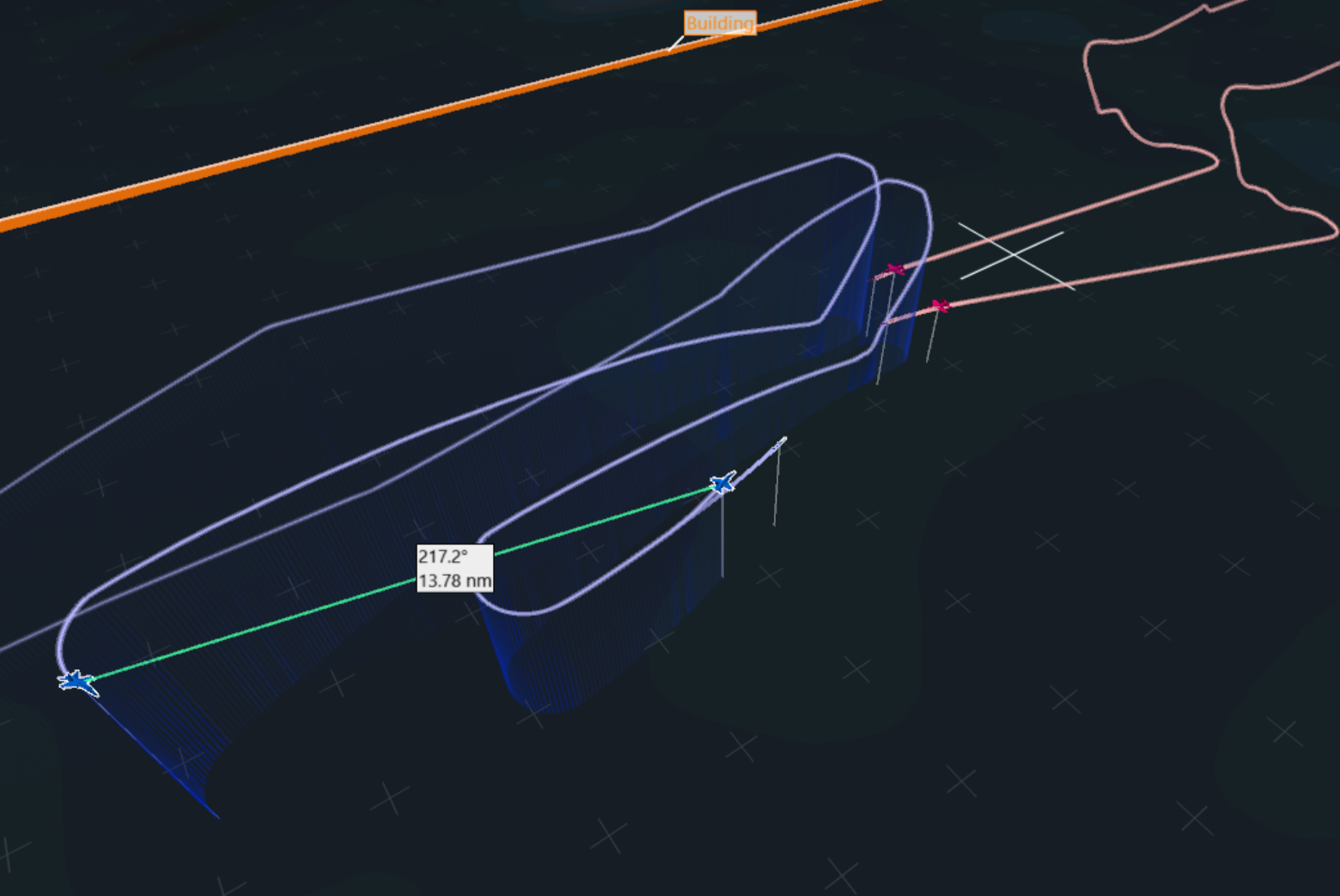}
      \caption{bait-and-flank tactic}
      \label{fig:bait-and-flank}
  \end{subfigure}
  \hfill
  \begin{subfigure}[b]{0.31\textwidth}
      \centering
      \includegraphics[width=\textwidth]{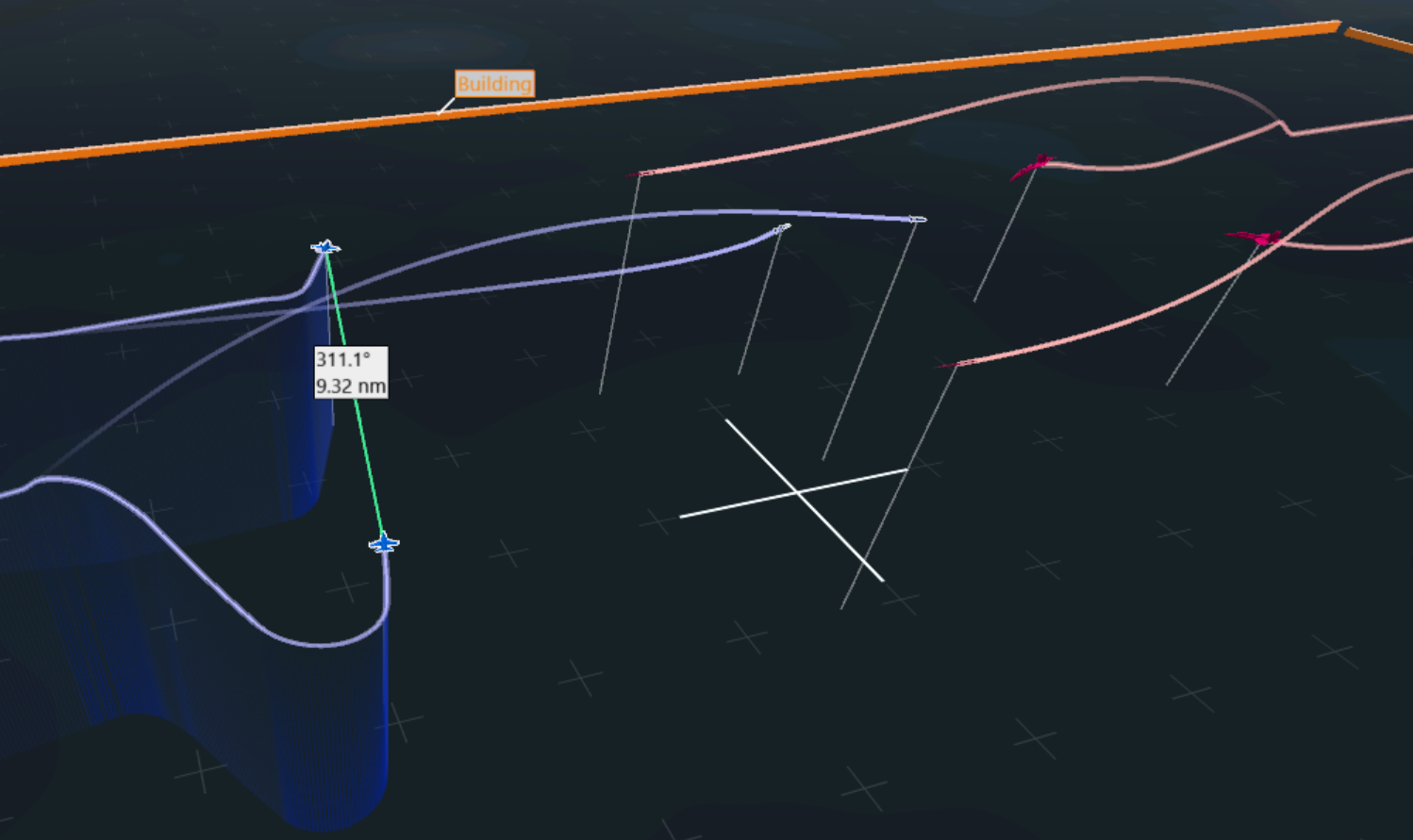}
      \caption{focus-fire tactic}
      \label{fig:focus-fire}
  \end{subfigure}
  \hfill
  \begin{subfigure}[b]{0.33\textwidth}
      \centering
      \includegraphics[width=\textwidth]{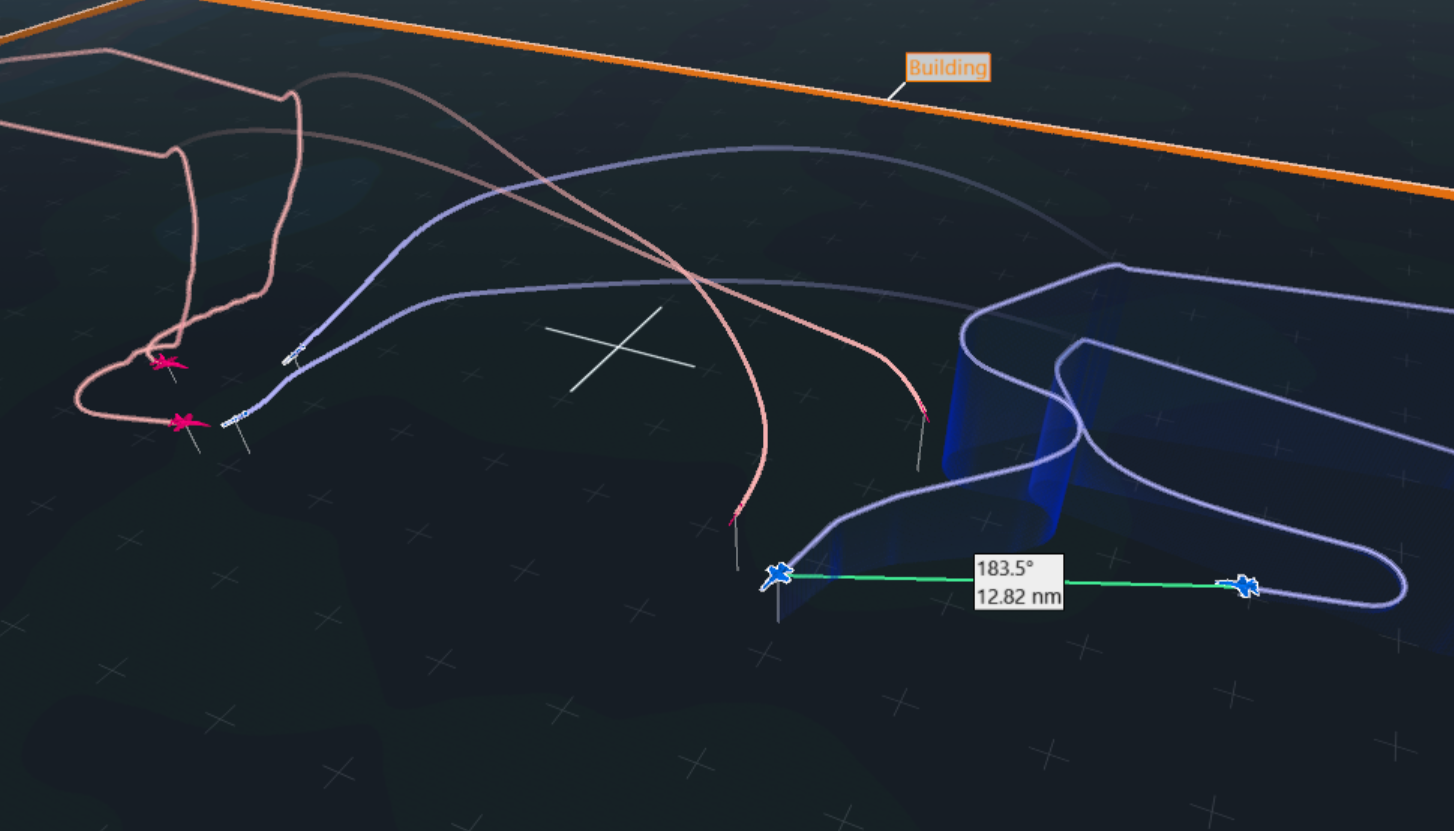}
      \caption{hit the target}
      \label{fig:Successfully-hit}
  \end{subfigure}
  \caption{Visualization of emergent tactical behaviors in a 2v2 engagement.}
  \label{fig:Visualization}
\end{figure*}

As shown in Fig.~\ref{fig:bait-and-flank}, asymmetric role assignment fosters an emergent bait-and-flank tactic. The Supporter deliberately maintains a forward trajectory to draw enemy fire, creating a tactical window for the Leader to execute a lateral flanking maneuver and establish a favorable off-axis attack geometry. 
Furthermore, Fig.~\ref{fig:focus-fire} illustrates a coordinated focus-fire behavior. After evading incoming threats, both agents converge on a single adversary guided by the target-priority auxiliary task, launching missiles from complementary angles to maximize the kill probability. Notably, as the threat diminishes, the Supporter dynamically transitions from a defensive to an offensive posture and successfully strikes the target (shown in Fig.~\ref{fig:Successfully-hit}), highlighting the temporal adaptability of our role commitment mechanism.

\section{Conclusion}
In this paper, we presented DRG-MAPPO, a hierarchical MARL framework designed to address coordination challenges in multi-UAV air combat. By integrating a graph-based relational encoder with a dynamic role assignment mechanism, our approach effectively decouples high-level tactical intent from low-level maneuver execution. Extensive simulations in a high-fidelity BVR environment demonstrate that DRG-MAPPO achieves a superior win rate of 87\%. Furthermore, the emergence of advanced collaborative tactics, such as bait-and-flank and focus-fire, underscores the effectiveness of our framework in facilitating complex multi-agent decision-making. Future work will explore the scalability of this hierarchical paradigm in larger-scale swarm confrontations.

\begin{credits}
\subsubsection{\ackname} 
This research received no external funding. The authors express gratitude to the anonymous reviewers for their insightful comments.

\subsubsection{\discintname}
The authors have no competing interests to declare that are relevant to the content of this article.

\end{credits}

%
%
%
%
\newpage
\bibliographystyle{splncs04}
\bibliography{references}






\end{document}